\documentclass{article}

\usepackage{times}
\usepackage{graphicx} 
\usepackage{subfigure} 

\usepackage{natbib}

\usepackage{algorithm}
\usepackage{algorithmic}

\usepackage{hyperref}

\usepackage[accepted]{icml2017}

\icmltitlerunning{Dance Dance Convolution}

\usepackage{amsmath}
\usepackage{booktabs}
\usepackage{float}

\begin{document} 

\twocolumn[
\icmltitle{Dance Dance Convolution}



\icmlsetsymbol{equal}{*}

\begin{icmlauthorlist}
\icmlauthor{Chris Donahue}{ucsdmus}
\icmlauthor{Zachary C. Lipton}{ucsdcs}
\icmlauthor{Julian McAuley}{ucsdcs}
\end{icmlauthorlist}

\icmlaffiliation{ucsdmus}{UCSD Department of Music, San Diego, CA}
\icmlaffiliation{ucsdcs}{UCSD Department of Computer Science, San Diego, CA}

\icmlcorrespondingauthor{Chris Donahue}{cdonahue@ucsd.edu}

\icmlkeywords{machine learning, ICML, music information retrieval, MIR, Dance Dance Revolution, DDR, procedural content generation, automatic level design, Dance Dance Convolution, DDC, recurrent neural networks, RNN, convolutional neural networks, CNN, LSTM}

\vskip 0.3in
]



\printAffiliationsAndNotice{}  

\begin{abstract} 
\emph{Dance Dance Revolution} (DDR)
is a popular \emph{rhythm-based} video game.
Players perform steps on a \emph{dance platform} 
in synchronization with music 
as directed by on-screen \emph{step charts}.
While many step charts are available in standardized packs,
players may grow tired of existing charts, 
or wish to dance to a song 
for which no chart exists.
We introduce the task of \emph{learning to choreograph}.
Given a raw audio track, the goal is to produce a new step chart.
This task decomposes naturally into two subtasks: 
deciding when to place steps 
and deciding which steps to select.
For the \emph{step placement} task,
we combine recurrent and convolutional neural networks to ingest spectrograms of low-level audio features to predict steps, conditioned on chart difficulty.
For \emph{step selection}, we present a conditional LSTM generative model that substantially outperforms $n$-gram and fixed-window approaches.
\end{abstract} 

\section{Introduction}
\label{sec:introduction}
\begin{figure}
  \centering
   \includegraphics[width=0.95\linewidth]{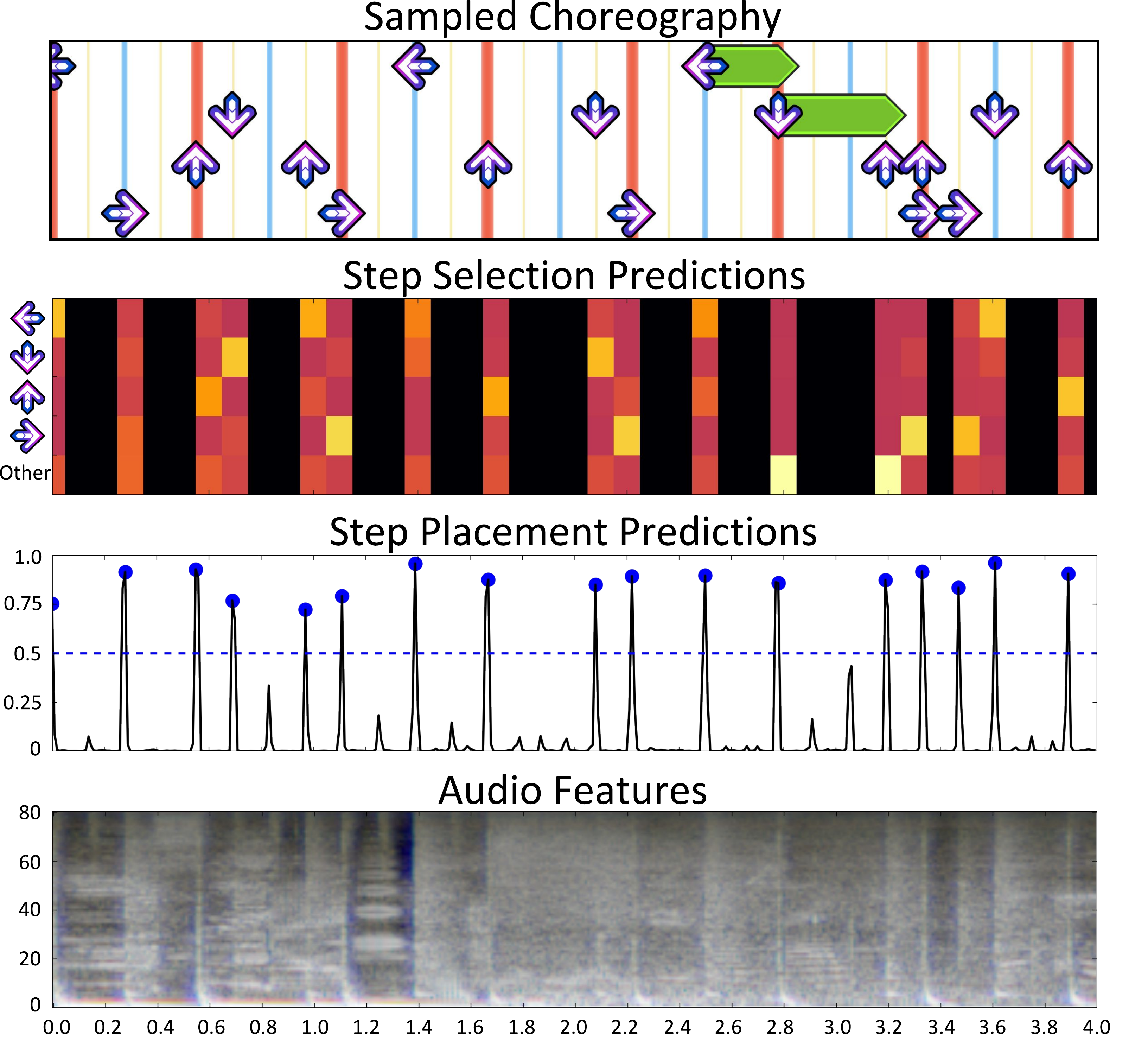}
   \caption{
Proposed \emph{learning to choreograph} pipeline 
for four seconds of the song \emph{Knife Party feat. Mistajam - Sleaze}.
The pipeline ingests audio features (\textbf{Bottom}) and produces a playable DDR choreography (\textbf{Top}) corresponding to the audio.
}
   \label{fig:sexyfig1}
\end{figure}

\emph{Dance Dance Revolution} (DDR) is a rhythm-based video game with millions of players worldwide~\citep{hoysniemi2006international}. Players perform steps atop a dance platform, 
following prompts from an on-screen \emph{step chart}
to step on the platform's buttons
at specific, musically salient points in time. 
A player's score depends upon both hitting the correct buttons and hitting them at the correct time.
Step charts vary in difficulty with
harder charts containing more steps 
and more complex sequences.
The dance pad contains \emph{up}, \emph{down}, \emph{left}, and \emph{right} arrows,
each of which can be in one of four states:
\emph{on}, \emph{off}, \emph{hold}, or \emph{release}.
Because the four arrows can be activated or released
independently, there are $256$ possible step combinations at any instant.

\vspace{3mm}

\pagebreak

\begin{figure}
  \centering
    \includegraphics[width=0.95\linewidth]{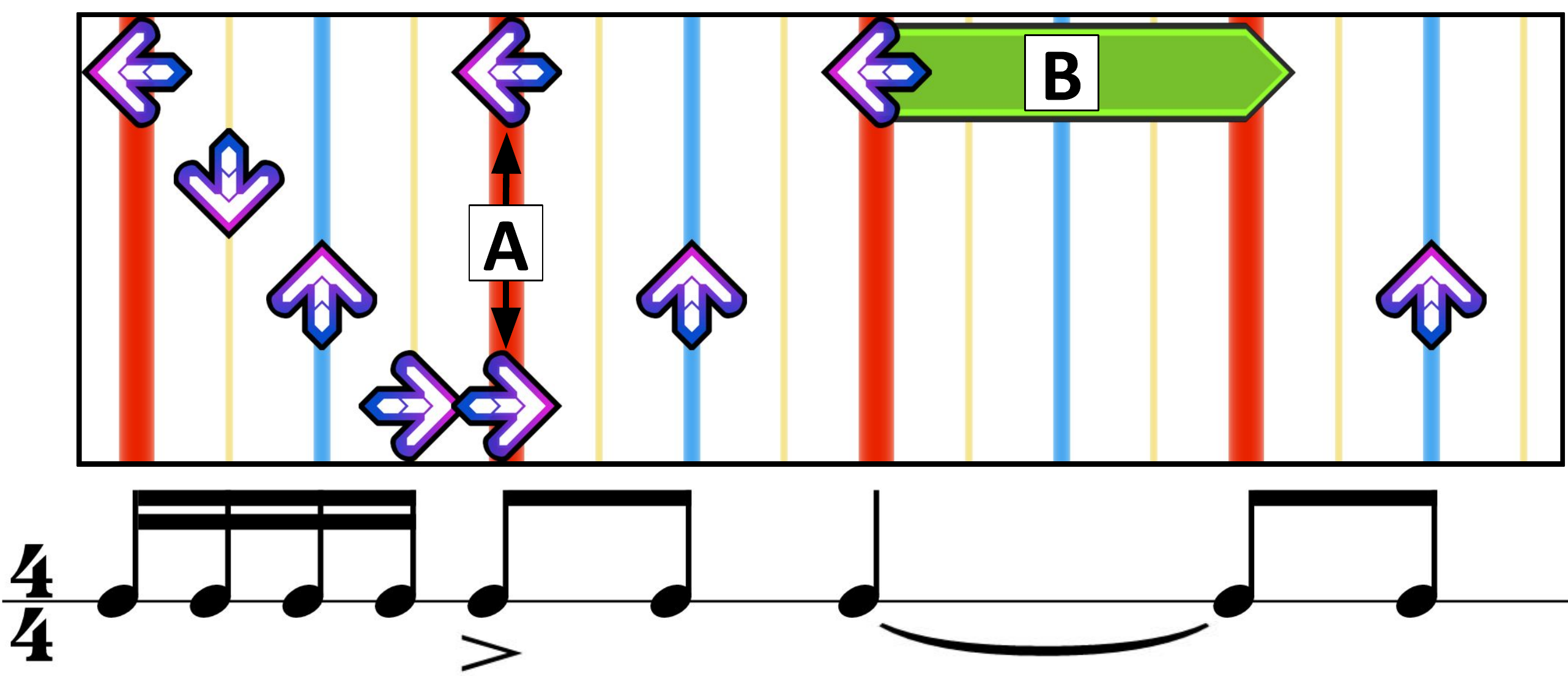}
    \caption{A four-beat measure of a typical chart and its rhythm depicted in musical notation. \textbf{Red:} quarter notes, \textbf{Blue:} eighth notes, \textbf{Yellow:} sixteenth notes, \textbf{(A)}: \emph{jump} step, 
\textbf{(B)}: \emph{freeze} step}
\label{fig:stepchart}
\end{figure}

Step charts exhibit rich structure and complex semantics
to
ensure that step sequences are both challenging and enjoyable.
Charts tend to mirror musical structure:
particular sequences of steps correspond to different motifs (Figure \ref{fig:stepchart}), 
and entire passages may reappear as sections of the song are repeated. 
Moreover, chart authors strive to 
avoid patterns that would compel a player to face away from the screen.

The DDR community uses simulators,
such as the open-source \emph{StepMania},
that allow fans to create and play their own charts.
A number of prolific authors produce and disseminate
\emph{packs} of charts,
bundling metadata with relevant recordings.
Typically, for each song, 
packs contain one chart for each of five difficulty levels.

Despite the game's popularity, 
players have some reasonable complaints:
For one,
packs are limited 
to songs with favorable licenses,
meaning players may be unable to
dance to their favorite songs.
Even when charts are available,
players may tire of repeatedly
performing the same charts.
Although players can produce their own charts,
the process is painstaking and requires significant expertise.

In this paper, we seek to automate the process of step chart generation 
so that players can dance to a wider variety of charts on any song of their choosing.
We introduce the task of \emph{learning to choreograph},
in which we learn to generate
step charts from raw audio.
Although this task has previously been approached via ad-hoc methods, we are the first to cast it as a learning task in which we seek to mimic the semantics of human-generated charts.
We break the problem into two subtasks:
First, \emph{step placement} consists of 
identifying a set of timestamps in the song 
at which to place steps.
This process can be conditioned on a player-specified difficulty level.
Second, \emph{step selection}
consists of choosing which steps to place at each timestamp. 
Running these two steps in sequence yields
a playable \emph{step chart}. 
This process is depicted 
in Figure \ref{fig:sexyfig1}.

Progress on learning to choreograph
may also lead to advances in music information retrieval (MIR).
Our step placement task, for example,
closely resembles \emph{onset detection},
a well-studied MIR problem.
The goal of onset detection is to identify the times of all
musically salient events,
such as melody notes or drum strikes.
While not every onset in our data corresponds to a DDR step, 
every DDR step corresponds to an onset.
In addition to marking steps, 
DDR packs specify a metronome click track for each song.
For songs with changing tempos, 
the exact location of each change and the new tempo are annotated. 
This click data could help to spur algorithmic innovation 
for \emph{beat tracking} and \emph{tempo detection}.

Unfortunately, MIR research is
stymied by the difficulty of accessing large, well-annotated datasets.
Songs are often subject to copyright issues, and thus must be gathered by each researcher independently. 
Collating audio with separately-distributed metadata is nontrivial and error-prone owing to
the multiple
available versions of many songs.
Researchers often must manually align their version of a song to the metadata.
In contrast, our dataset is publicly available, standardized
and contains
meticulously-annotated labels as well as the relevant recordings.

We believe that DDR charts represent an abundant and under-recognized source of annotated data for MIR research.
\emph{StepMania Online},
a popular repository of DDR data, 
distributes over $350\mathit{Gb}$ of packs with annotations for more than $100\mathit{k}$ songs. 
In addition to introducing a novel task and methodology, 
we contribute two large public datasets, which we consider to be of notably high quality and consistency.\footnote{\label{ddcrepo}\textbf{\texttt{https://github.com/chrisdonahue/ddc}}}
Each dataset is a collection of recordings and step charts.
One contains charts by a single author
and the other by multiple authors.

For both prediction stages 
of learning to choreograph,
we demonstrate the superior performance of neural networks over strong alternatives.
Our best model for step placement
jointly learns a convolutional neural network (CNN) representation
and a recurrent neural network (RNN), 
which integrates information across consecutive time slices.
This method outperforms CNNs alone, multilayer perceptrons (MLPs), 
and linear models. 

Our best-performing system for step selection consists 
of a conditional LSTM generative model.
As auxiliary information,
the model takes \emph{beat phase},
a number representing
the fraction of a beat at which a step occurs.
Additionally, the best models receive
the time difference (measured in beats) since the last and until the next step.
This model selects steps that are more consistent with expert authors than the best $n$-gram and fixed-window models,
as measured by perplexity and per-token accuracy.

\subsection{Contributions}
In short, our paper offers the following contributions:
\begin{itemize}
	\item We define \emph{learning to choreograph}, a new task with real-world usefulness and strong connections to fundamental problems in MIR.   
    \item We introduce two large, curated datasets for benchmarking DDR choreography algorithms. 
    They represent an under-recognized source of music annotations.
	\item We introduce an effective pipeline for learning to choreograph with deep neural networks.\footnote{Demonstration showing human choreography alongside output of our method:
\textbf{\texttt{https://youtu.be/yUc3O237p9M}}
}
\end{itemize}

\section{Data}
\label{sec:data}
\begin{figure}[t!]
   \centering
     \includegraphics[width=0.95\linewidth]{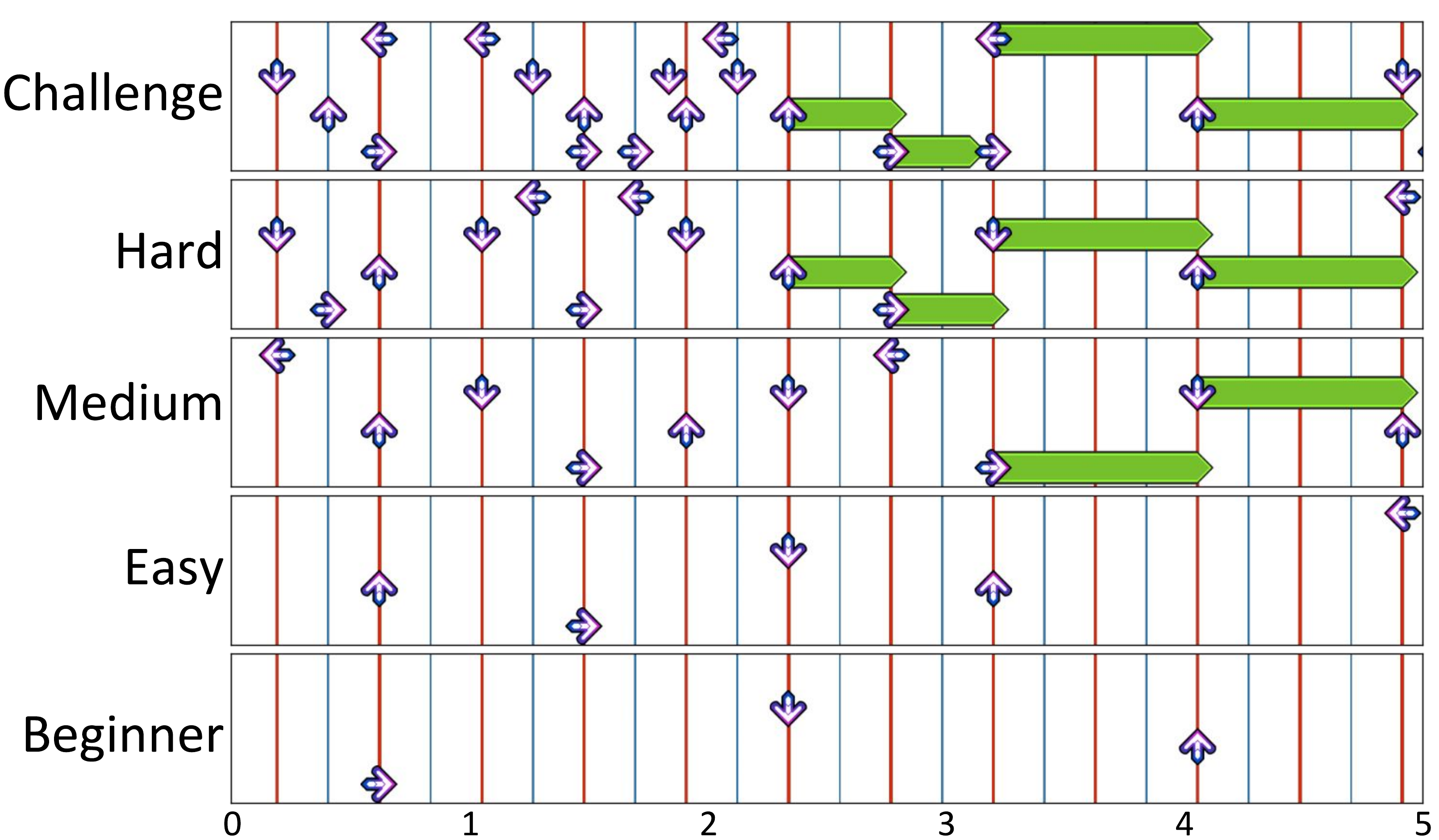}
     \caption{Five seconds of choreography by difficulty level for the song \emph{KOAN Sound - The Edge} from the Fraxtil training set.}
       \label{fig:fraxtil_diff}
\end{figure}

Basic statistics of our two datasets 
are shown in
Table~\ref{tab:data_stats}.
The first dataset contains $90$ songs choreographed by a single prolific author who works under the name \emph{Fraxtil}.
This dataset contains five charts per song
corresponding to increasing difficulty levels.
We find that while these charts overlap significantly, 
the lower difficulty charts 
are not strict subsets of the higher difficulty charts (Figure \ref{fig:fraxtil_diff}). 
The second dataset is a larger, multi-author collection 
called \emph{In The Groove} (ITG); 
this dataset contains $133$ songs 
with one chart per difficulty, 
except for $13$ songs that lack charts 
for the highest difficulty. 
Both datasets contain electronic music with constant tempo and a strong beat, characteristic of music favored by the DDR community.

\begin{table}[hbt!]
\centering
\caption{Dataset statistics}
\footnotesize
\begin{tabular}{l|rr}
\toprule 
Dataset  &  \textbf{\textbf{Fraxtil}} & \textbf{\textbf{ITG}} \\
\midrule
Num authors & 1 & 8 \\
Num packs & 3 & 2 \\
Num songs & 90 (3.1 hrs) & 133 (3.9 hrs)\\
Num charts & 450 (15.3 hrs) & 652 (19.0 hrs) \\
Steps/sec & 3.135 & 2.584 \\
Vocab size & 81 & 88 \\
\bottomrule
\end{tabular} 
\label{tab:data_stats}
\end{table}

Note that while the total number of songs is relatively small, 
when considering all charts across all songs
the datasets contain around 35 hours of annotations and 350,000 steps.
The two datasets have similar vocabulary sizes ($81$ and $88$ distinct step
combinations, respectively).
Around $84\%$ of the steps in both datasets consist of a single,
instantaneous
arrow.

Step charts contain several invariances, for example interchanging all instances of left and right results in an equally plausible sequence of steps. To augment the amount of data available for training, we generate four instances of each chart, by mirroring left/right, up/down (or both). Doing so considerably improves performance in practice.

In addition to encoded audio, packs consist of metadata
including a song's title, artist, a list of time-stamped tempo changes, 
and a time offset to align the recording to the tempos.
They also contain information such as the chart difficulties and the name of the choreographer.
Finally, the metadata contains a full list of steps, 
marking the measure and beat of each.
To make this data easier to work with, 
we convert it to a canonical form consisting
of (\emph{beat}, \emph{time}, \emph{step}) tuples.

\begin{figure}
  \centering
    \includegraphics[width=0.95\linewidth]{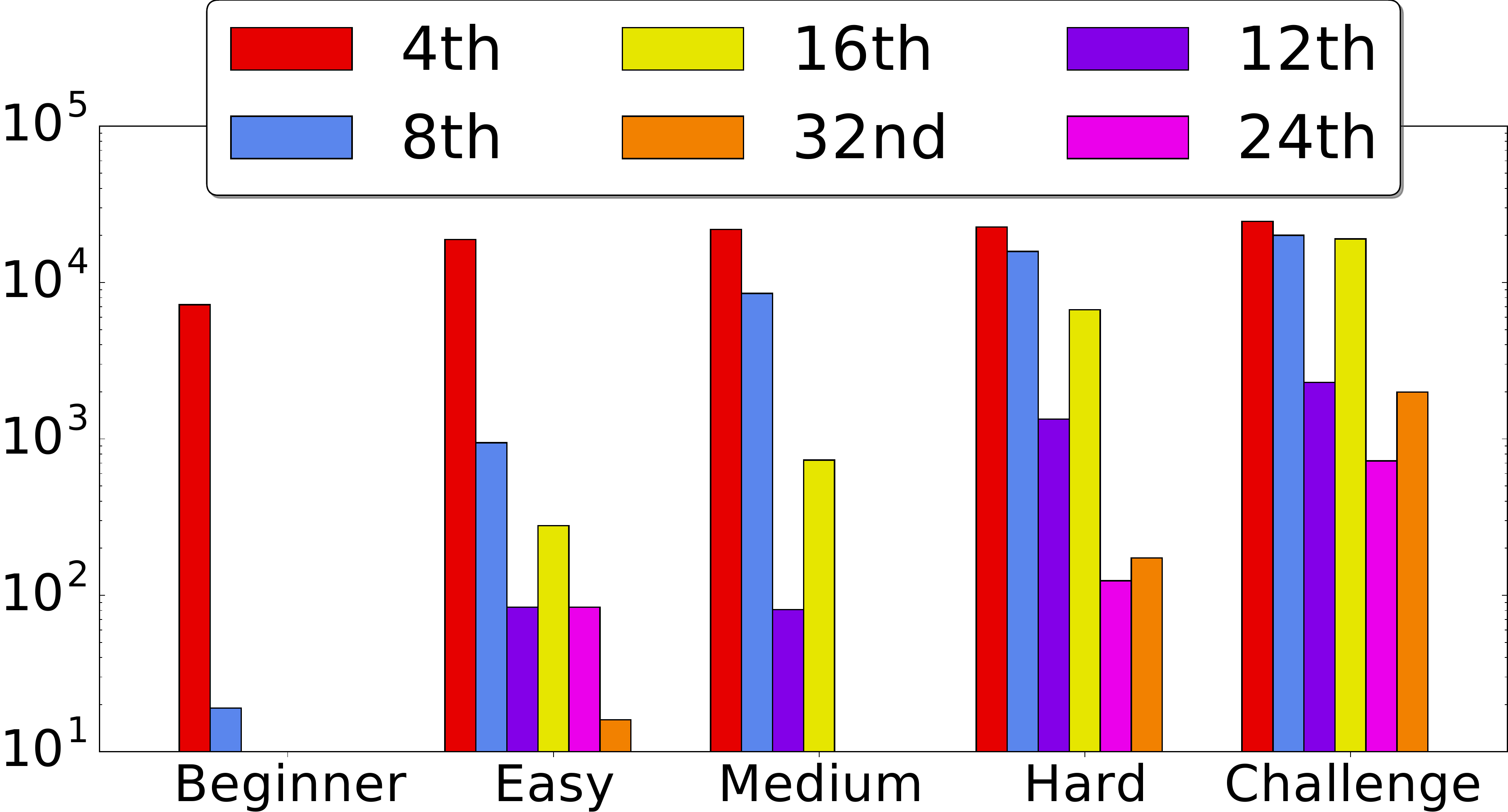}
    \caption{Number of steps per rhythmic subdivision by difficulty in the Fraxtil dataset.}
      \label{fig:fraxtil_subdivs}
\end{figure}

The charts in both datasets
echo high-level rhythmic structure in the music.
An increase in difficulty corresponds to
increasing propensity for steps to appear at finer rhythmic subdivisions.
Beginner charts tend to contain 
only quarter notes and eighth notes. 
Higher-difficulty charts reflect
more complex rhythmic details in the music,
featuring higher densities of eighth and sixteenth note steps (8th, 16th) 
as well as triplet patterns (12th, 24th)
(Figure \ref{fig:fraxtil_subdivs}).

\section{Problem Definition}
\label{sec:problem}
A step can occur in up to $192$ different locations (subdivisions) within each measure.
However, measures contain roughly 6 steps on average.
This level of sparsity makes it difficult to uncover patterns 
across  long sequences of (mostly empty) frames 
via a single end-to-end sequential model.
So, to make automatic DDR choreography tractable,
we decompose it into two subtasks: 
step placement and step selection.

In \emph{step placement}, our goal is to decide
at what precise times to place steps. 
A step placement algorithm 
ingests raw audio features
and outputs timestamps corresponding to steps.
In addition to the audio signal, 
we provide step placement algorithms 
with a one-hot representation of the intended difficulty rating for the chart.

\emph{Step selection} involves taking a discretized list of step times computed during step placement and mapping each of these 
to a DDR step.
Our approach to this problem
involves modeling the probability distribution $P(m_n | m_1, \ldots, m_{n-1})$ where $m_n$ is the $n^{\text{th}}$ step in the sequence.
Some steps 
require that the player hit two or more
arrows
at once, a \emph{jump}; or hold on one
arrow
for some duration, a \emph{freeze} (Figure \ref{fig:stepchart}).

\section{Methods}
\label{sec:methods}
We now describe our specific solutions to the step placement and selection problems.
Our basic pipeline works as follows:
(1)~extract an audio feature representation;
(2)~feed this
representation into a step placement algorithm, 
which estimates probabilities that
a ground truth step lies within that frame;
(3)~use a peak-picking process on this sequence of probabilities to
identify the precise timestamps at which to place steps; 
and finally (4)~given a sequence of timestamps,
use a step selection algorithm 
to choose which steps to place at each time. 

\subsection{Audio Representation}
Music files arrive 
as lossy encodings at $44.1 \mathit{kHz}$. 
We decode the audio files into stereo PCM audio
and average the two channels 
to produce a monophonic representation. 
We then compute a multiple-timescale short-time Fourier transform (STFT)
using window lengths of $23 \mathit{ms}$, $46 \mathit{ms}$, and $93 \mathit{ms}$ 
and a stride of $10 \mathit{ms}$.
Shorter window sizes preserve low-level features 
such as pitch and timbre
while larger window sizes provide more context
for high-level features such as melody and rhythm~\citep{hamel2012building}.

Using the ESSENTIA library~\citep{bogdanov2013essentia}, we reduce the dimensionality of the STFT magnitude spectra to $80$ frequency bands
by applying a Mel-scale~\citep{stevens1937scale} filterbank. 
We scale the filter outputs logarithmically 
to better represent human perception of loudness. 
Finally,
we prepend and append seven frames of past and future context respectively to each frame. 

For fixed-width methods,
the final audio representation is a $15\times80\times3$ tensor.
These correspond to the temporal width of $15$
representing $150 \mathit{ms}$ of audio context, 
$80$ frequency bands, and
$3$ different window lengths.
To better condition the data for learning, 
we normalize each frequency band to zero mean and unit variance. 
Our approach to acoustic feature representation closely follows the work of \citet{schluter2014improved},
who develop similar representations 
to perform onset detection with CNNs.

\vspace{2.2mm}

\subsection{Step Placement}\label{sec:placement_methods}

\begin{figure}
  \centering
  \includegraphics[width=0.75\linewidth]{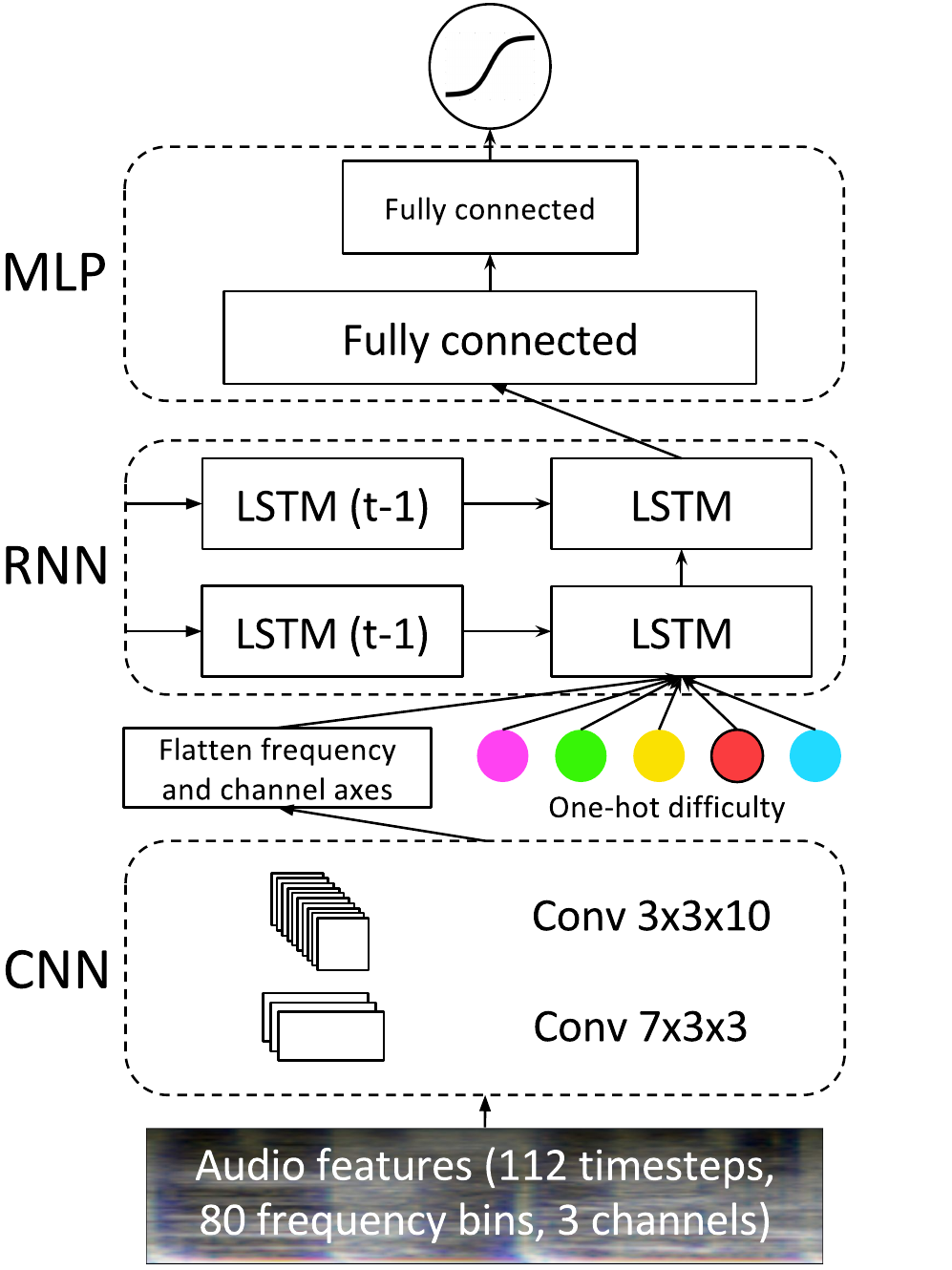}
  \caption{C-LSTM model used for step placement}
  \label{fig:placement_net}
\end{figure}

We consider several 
models to address the step placement task. 
Each model's output consists of a single sigmoid unit
which estimates the probability that a step is placed.
For all models, we augment the audio features 
with a one-hot representation of difficulty.

Following state-of-the-art work on onset detection \cite{schluter2014improved}, 
we adopt a convolutional neural network (CNN) 
architecture.
This model consists of two convolutional layers followed by two fully connected layers.
Our first convolutional layer has 10 filter kernels that are 7-wide in time and 3-wide in frequency.
The second layer has 20 filter kernels that are 3-wide in time and 3-wide in frequency. 
We apply 1D max-pooling after each convolutional layer, 
only in the frequency dimension,
with a width and stride of 3. Both convolutional layers use rectified linear units (ReLU)~\citep{glorot2011deep}.
Following the convolutional layers, 
we add two fully connected layers 
with rectifier activation functions
and $256$ and $128$ nodes, respectively.

To improve upon the CNN, 
we propose a C-LSTM model,
combining a convolutional encoding with an RNN that integrates information across longer windows of time. 
To encode the raw audio at each time step,
we first apply two convolutional layers (of the same shape as the CNN)
across the full unrolling length.
The output of the second convolutional layer is a $3$D tensor, 
which we flatten along the channel and frequency axes 
(preserving the temporal dimension). 
The flattened features at each time step
then become the inputs to a two-layer RNN.

The C-LSTM contains long short-term memory (LSTM) units~\citep{hochreiter1997long} with forget gates \citep{gers2000count}.
The LSTM consists of $2$ layers with $200$ nodes each.
Following the LSTM layers, we apply two fully connected ReLU layers of dimension $256$ and $128$. This architecture is depicted in Figure~\ref{fig:placement_net}.
We train this model using $100$ unrollings for backpropagation through time.

A chart's intended difficulty
influences decisions both about how many steps to place and where to place them. 
For low-difficulty charts, 
the average number of steps per second is less than one.
In contrast, the highest-difficulty charts 
exceed seven steps per second. 
We trained all models both with and without conditioning on difficulty,
and found the inclusion of this feature to be informative.
We concatenate difficulty features to the flattened output of the CNN before feeding the vector to the fully connected (or LSTM) layers (Figure \ref{fig:placement_net}).\footnote{For LogReg and MLP, 
we add difficulty to input layer.}
We initialize weight matrices following the scheme of~\citet{glorot2010understanding}.

\paragraph{Training Methodology}
We minimize binary cross-entropy
with mini-batch stochastic gradient descent.
For all models we train with batches of size $256$, 
scaling down gradients when their 
$l_{2}$ norm exceeds $5$.
We apply $50\%$ dropout following each LSTM and fully connected layer.
For LSTM layers, we apply dropout in the input to output 
but not temporal directions, 
following best practices from \citep{zaremba2014recurrent,lipton2015learning, dai2015semi}.
Although the problem exhibits pronounced class imbalance  (97\% negatives), 
we achieved better results training on imbalanced data than with re-balancing schemes.
We exclude all examples before the first step in the chart or after the last step as charts typically do not span the entire duration of the song.

For recurrent neural networks, 
the target at each frame is the ground truth value corresponding to that frame. 
We calculate updates using backpropagation through time with $100$ steps of unrolling, 
equal to one second of audio or two beats on a typical track ($120$ BPM). 
We train all networks 
with early-stopping determined by 
the area under the precision-recall curve on validation data. 
All models satisfy this criteria within $12$ hours of training on a single machine with an NVIDIA Tesla K40m GPU.

\subsection{Peak Picking}

\begin{figure}[t]
  \centering
  \includegraphics[width=0.95\linewidth]{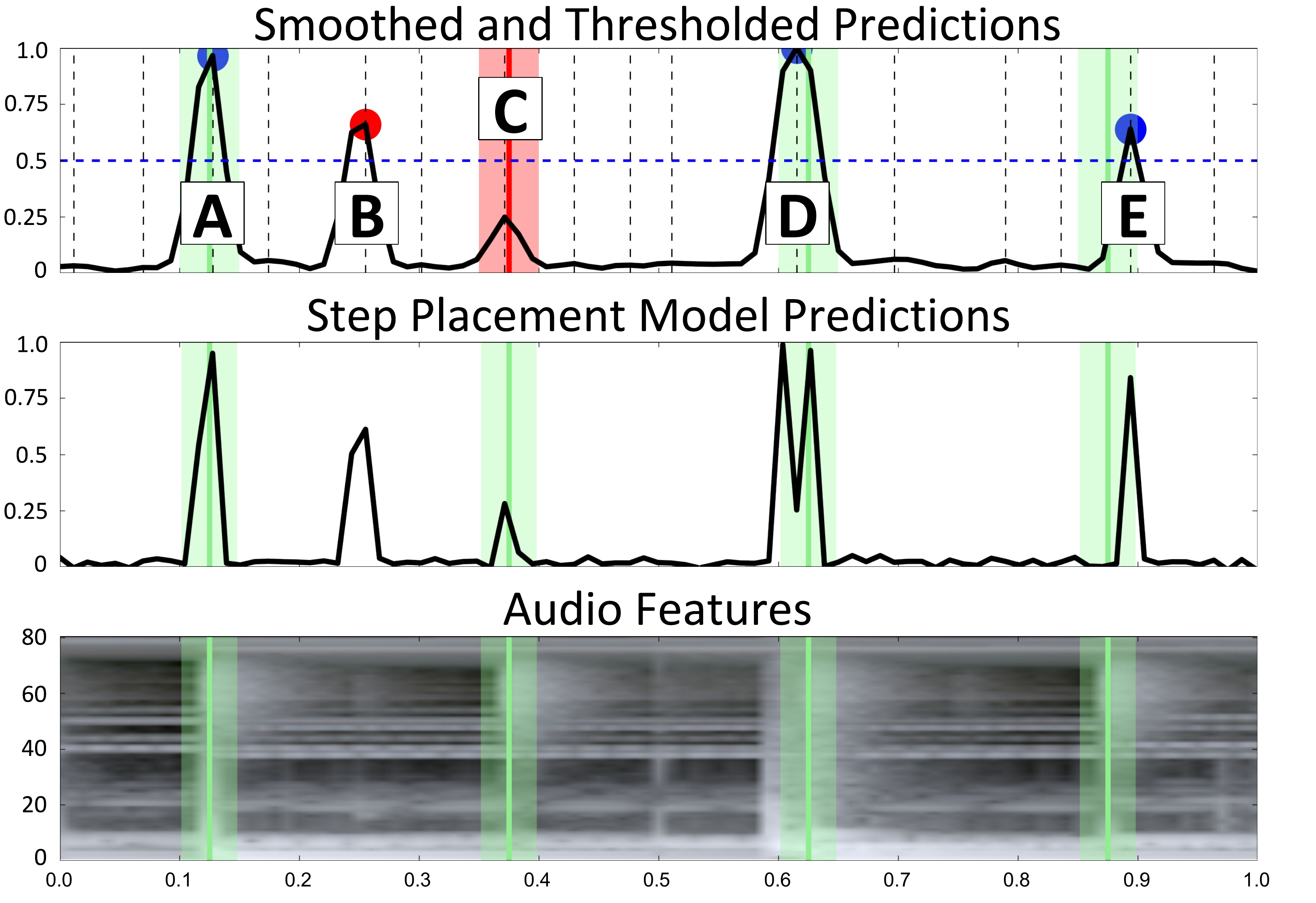}
  \caption{One second of peak picking. \textbf{Green: } Ground truth region \textbf{(A)}: true positive, \textbf{(B)}: false positive, \textbf{(C)}: false negative, \textbf{(D)}: two peaks smoothed to one by Hamming window, \textbf{(E)}: misaligned peak accepted as true positive by $\pm 20\mathit{ms}$ tolerance}
  \label{fig:peak_pick}
\end{figure}

Following standard practice for onset detection, 
we convert sequences of step probabilities 
into a discrete set of chosen placements via a peak-picking process. 
First we run our step placement algorithm over an entire song to assign the probabilities of a step occurring within each $10\mathit{ms}$ frame.\footnote{In DDR, 
scores depend on the accuracy of a player's step timing.
The highest scores require that a step is performed within $22.5\mathit{ms}$ of its appointed time;
this suggests that a reasonable algorithm 
should place steps with an even finer level of granularity.}
We then convolve this sequence of predicted probabilities with a Hamming window,
smoothing the predictions and suppressing double-peaks from occurring within a short distance.
Finally, we apply a constant threshold to choose which peaks are high enough (Figure \ref{fig:peak_pick}).
Because the number of peaks varies according to chart difficulty, we choose a different threshold per difficulty level.
We consider predicted placements to be true positives if they lie within a $\pm 20\mathit{ms}$ window of a ground truth.

\subsection{Step Selection}\label{sec:selection_methods}

\begin{figure}[t]
  \centering
  \includegraphics[width=0.95\linewidth]{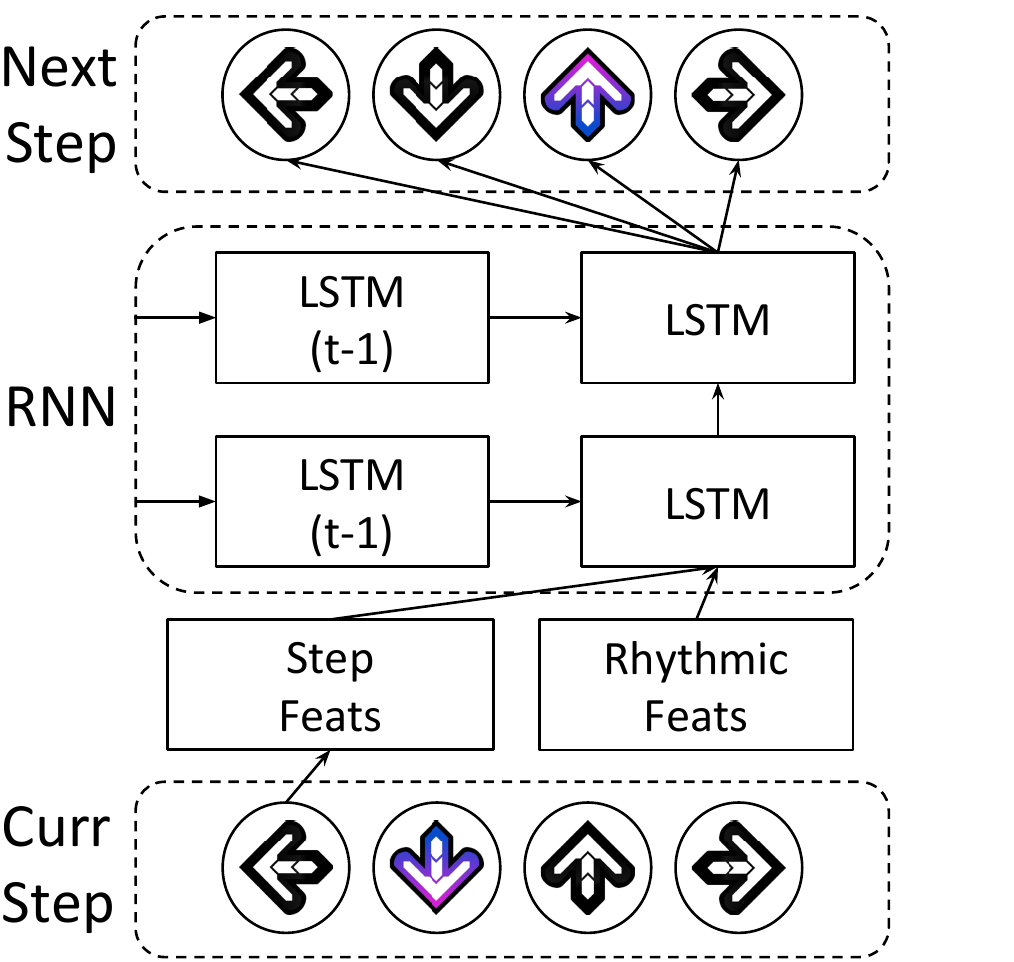}
  \caption{LSTM model used for step selection}
  \label{fig:selection_net}
\end{figure}

We treat the step selection task as a sequence generation problem.
Our approach follows related work in language modeling
where RNNs are well-known to produce coherent
text that captures long-range
relationships~\citep{mikolov2010recurrent, sutskever2011generating, sundermeyer2012lstm}. 

Our LSTM model passes over the ground truth step placements and predicts the next token given the previous sequence of tokens.
The output is a softmax distribution over the $256$ possible steps.
As input, we use a more compact \emph{bag-of-arrows} representation
containing $16$ features ($4$ per arrow) to depict the previous step.
For each arrow, the $4$ corresponding features 
represent the states \emph{on, off, hold,} and \emph{release}.
We found the bag-of-arrows to give equivalent performance to the one-hot representation while requiring fewer parameters.
We add an additional feature 
that functions as a \emph{start} token
to denote the first step of a chart.
For this task, we use an LSTM with $2$ layers of $128$ cells each.

Finally, we provide additional musical context to the step selection models by conditioning on rhythmic features
(Figure \ref{fig:selection_net}).
To inform models of the non-uniform spacing of the 
step placements, we consider the following three features:
(1)~\emph{$\Delta$-time} adds two features representing the time since the previous step and the time until the next step;
(2)~\emph{$\Delta$-beat} adds two features representing the number of beats since the previous and until the next step;
(3)~\emph{beat phase} adds four features representing which 16th note subdivision of the beat the current step most closely aligns to.

\paragraph{Training Methodology}
For all neural network models, 
we learn parameters by minimizing cross-entropy.
We train with mini-batches of size $64$,
and scale gradients using the same scheme as for step placement. 
We use $50\%$ dropout during training for both the MLP and RNN models in the same fashion as for step placement.
We use $64$ steps of unrolling, representing an average of $100$ seconds for the easiest charts and $9$ seconds for the hardest. 
We apply early-stopping determined by average per-step cross entropy on validation data. 
All models satisfy this criteria within $6$ hours of training on a single machine with an NVIDIA Tesla K40m GPU.

\section{Experiments}
\label{sec:experiments}
For both the Fraxtil and ITG datasets we apply $80\%$, $10\%$, $10\%$
splits for training, validation, and test data, respectively.
Because of correlation between charts for the same song of varying difficulty, 
we ensure that all charts for a particular song 
are grouped together in the same split.

\subsection{Step Placement}

\begin{table}[t!]
\centering
\caption{Results for step placement experiments}
\footnotesize
\begin{tabular}{ll|cccc}
\toprule
\textbf{\textbf{Model}} &  \textbf{Dataset} & \textbf{PPL} & \textbf{AUC} 
& $\!\!$\textbf{F-score$^{c}$}$\!\!\!$ & $\!\!$\textbf{F-score$^{m}$}$\!\!\!$ \\
\midrule
LogReg & Fraxtil & 1.205 & 0.601 & 0.609 & 0.667 \\
MLP & Fraxtil & 1.097 & 0.659 & 0.665 & 0.726 \\
CNN & Fraxtil & 1.082 & 0.671 & 0.678 & 0.750 \\
C-LSTM & Fraxtil & \textbf{1.070} & \textbf{0.682} & \textbf{0.681} & \textbf{0.756} \\
\midrule
LogReg & ITG & 1.123 & 0.599 & 0.634 & 0.652 \\
MLP & ITG & 1.090 & 0.637 & 0.671 & 0.704 \\
CNN & ITG & 1.083 & 0.677 & 0.689 & 0.719 \\
C-LSTM & ITG & \textbf{1.072} & \textbf{0.680} & \textbf{0.697} & \textbf{0.721} \\
\bottomrule
\end{tabular} 
\label{tab:steppla_res}
\end{table}

\begin{table}[t!]
\centering
\caption{Results for step selection experiments}
\footnotesize
\setlength{\tabcolsep}{4pt}
\begin{tabular}{lc|cc}
\toprule
\textbf{\textbf{Model}} &  \textbf{Dataset} & \textbf{PPL} & \textbf{Accuracy} \\
\midrule
KN5 & Fraxtil & 3.681 & 0.528 \\
MLP5 & Fraxtil & 3.744 & 0.543 \\
MLP5 + $\Delta$-time & Fraxtil & 3.495 & 0.553 \\
MLP5 + $\Delta$-beat + beat phase & Fraxtil & 3.428 & 0.557 \\
LSTM5 & Fraxtil & 3.583 & 0.558 \\
LSTM5 + $\Delta$-time & Fraxtil & 3.188 & 0.584 \\
LSTM5 + $\Delta$-beat + beat phase & Fraxtil & 3.185 & 0.581 \\
LSTM64 & Fraxtil & 3.352 & 0.578 \\
LSTM64 + $\Delta$-time & Fraxtil  & 3.107 & 0.606 \\
LSTM64 + $\Delta$-beat + beat phase & Fraxtil  & \textbf{3.011} & \textbf{0.613} \\
\midrule
KN5 & ITG & 5.847 & 0.356 \\
MLP5 & ITG & 5.312 & 0.376 \\
MLP5 + $\Delta$-time & ITG & 4.792 & 0.402 \\
MLP5 + $\Delta$-beat + beat phase & ITG & 4.786 & 0.401 \\
LSTM5 & ITG & 5.040 & 0.407 \\
LSTM5 + $\Delta$-time & ITG & 4.412 & 0.439 \\
LSTM5 + $\Delta$-beat + beat phase & ITG & 4.447 & 0.441 \\
LSTM64 & ITG & 4.780 & 0.426 \\
LSTM64 + $\Delta$-time & ITG  & \textbf{4.284} & \textbf{0.454} \\
LSTM64 + $\Delta$-beat + beat phase & ITG  & 4.342 & 0.444 \\
\bottomrule
\end{tabular} 
\label{tab:stepsel_res}
\end{table}

We evaluate the performance of our step placement methods against baselines via the methodology outlined below.

\paragraph{Baselines}
To establish reasonable baselines for step placement,
we first report the results of a logistic regressor (LogReg) 
trained on flattened audio features. 
We also report the performance of an MLP.
Our MLP architecture contains two fully-connected layers 
of size 256 and 128,
with rectifier nonlinearity applied to each layer.
We apply dropout with probability $50\%$ after each fully-connected layer during training. 
We model our CNN baseline 
on the method of \citet{schluter2014improved}, 
a state-of-the-art algorithm for onset detection.

\paragraph{Metrics}
We report each model's perplexity (PPL) 
averaged across each frame in each chart in the test data.
Using the sparse step placements, we calculate the average per-chart area under the precision-recall curve (AUC).
We average the best per-chart F-scores 
and report this value as F-score$^{c}$. 
We calculate the micro F-score across all charts 
and report this value as F-score$^{m}$.

In Table \ref{tab:steppla_res}, we list the results of our experiments for step placement.
For ITG, models were conditioned on not just difficulty but also a one-hot representation of chart author.
For both datasets, the C-LSTM model performs the best by all evaluation metrics.
Our models achieve significantly higher F-scores for harder difficulty step charts.
On the Fraxtil dataset, the C-LSTM achieves an F-score$^{c}$ of $0.844$ 
for the hardest difficulty charts
but only $0.389$ 
for the lowest difficulty.
The difficult charts contribute more to F-score$^{m}$ calculations 
because they have more ground truth positives.
We discuss these results further in Section \ref{sec:discussion}.

\subsection{Step Selection}\label{exp_step_sel}

\paragraph{Baselines}
For step selection,
we compare the performance of the conditional LSTM to an $n$-gram model.
Note that perplexity can be unbounded when a test set token 
is assigned probability $0$ by the generative model. 
To protect the $n$-gram models against unbounded loss on previously unseen $n$-grams, 
we use modified Kneser-Ney smoothing \citep{chen1998empirical},
following best practices in language modeling
\citep{mikolov2010recurrent,sutskever2011generating}.
Specifically, we train a smoothed $5$-gram model with backoff (\emph{KN5}) 
as implemented in \citet{stolcke2002srilm}.

Following the work of \citet{bengio2003neural}
we also compare against a fixed-window $5$-gram MLP 
which takes $4$ bag-of-arrows-encoded steps as input and predicts the next step.
The MLP contains two fully-connected layers with $256$ and $128$ nodes and $50\%$ dropout after each layer during training.
As with the LSTM, we train the MLP both with and without access to side features.
In addition to the LSTM with 64 steps of unrolling, 
we train an LSTM with 5 steps of unrolling.
These baselines show that the LSTM learns complex, long-range dependencies.
They also demonstrate the discriminative information conferred by the $\Delta$-time,
$\Delta$-beat, and beat phase features.

\paragraph{Metrics}
We report the average per-step perplexity,
averaging scores calculated separately on each chart.
We also report a per-token accuracy.
We calculate accuracy by comparing the ground-truth step to the argmax over a model's predictive distribution 
given the previous sequence of ground-truth tokens.
For a given chart, the per token accuracy 
is averaged across time steps.
We produce final numbers by averaging scores across charts.

In Table \ref{tab:stepsel_res} we present results for the step selection task.
For the Fraxtil dataset, 
the best performing model was the LSTM conditioned on both $\Delta$-beat and beat phase, while for ITG it was the LSTM conditioned on $\Delta$-time.
While conditioning on rhythm features was generally beneficial, the benefits of various features were not strictly additive. 
Representing 
$\Delta$-beat and $\Delta$-time as real numbers outperformed bucketed representations.

Additionally, we explored the possibility of incorporating more comprehensive representations of the audio into the step selection model.
We considered a variety of representations, 
such as conditioning on
CNN
features learned from the step placement task.
We also experimented with jointly learning a
CNN
audio encoder.
In all cases, these approaches led to rapid overfitting and never approached the performance of the conditional LSTM generative model; perhaps a much larger dataset could support these approaches.
Finally, we tried conditioning the step selection models on both difficulty and chart author but found these models to overfit quickly as well.

\section{Discussion}
\label{sec:discussion}
Our experiments establish the feasibility of using machine learning to automatically generate high-quality DDR charts from raw audio.
Our performance evaluations on both subtasks demonstrate 
the advantage of deep neural networks over classical approaches.
For step placement, the best performing model is an LSTM with CNN encoder, 
an approach which has been
used
for speech recognition \citep{amodei2015deep}, 
but, to our knowledge, never for music-related tasks. 
We noticed that by all metrics, 
our models perform better 
on higher-difficulty charts.
Likely, this owes to the comparative class imbalance of the lower difficulty charts.

\begin{figure}
  \centering
  \includegraphics[width=0.98\linewidth]{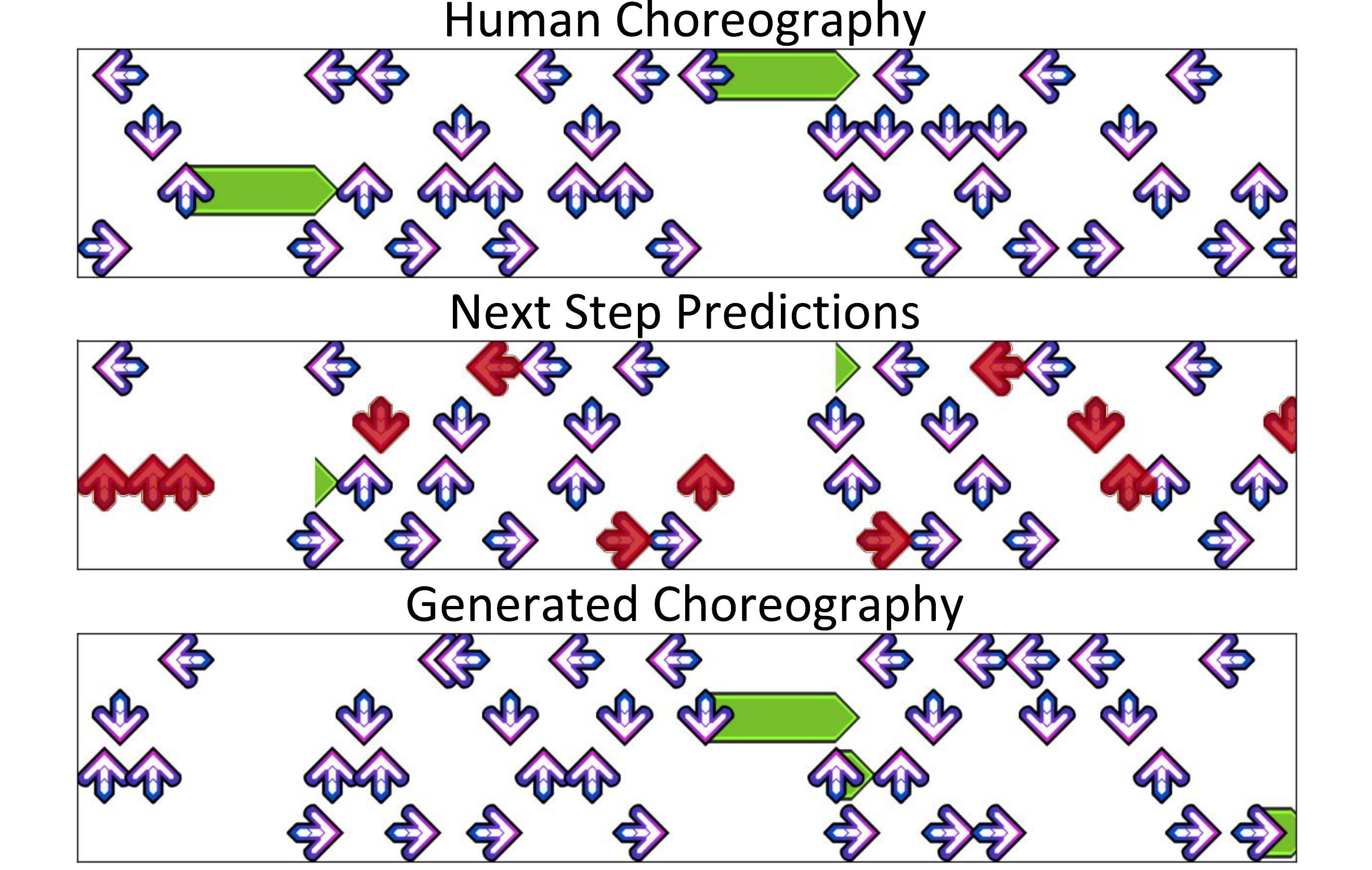}
  \caption{\textbf{Top: } A real step chart from the \emph{Fraxtil} dataset on the song \emph{Anamanaguchi - Mess}. \textbf{Middle:} One-step lookahead \emph{predictions} for the LSTM model, given Fraxtil's choreography as input. The model predicts the next step with high accuracy (errors in red). \textbf{Bottom: } Choreography generated by conditional LSTM model.}
  \label{fig:stepsel_gen}
\end{figure}

The superior performance of LSTMs over fixed-window approaches on step selection
suggests both that DDR charts exhibit long range dependencies and that recurrent neural networks can exploit this complex structure. 
In addition to reporting quantitative results, 
we visualize the step selection model's next-step predictions.
Here, we give the entire ground truth sequence as input 
but show the predicted next step at each time. 
We also visualize a generated choreography, where each sampled output from the LSTM is fed in as the subsequent input (Figure \ref{fig:stepsel_gen}).
We note the high accuracy of the model's predictions
and qualitative similarity of the generated sequence to Fraxtil's choreography.

For step selection, we notice that modeling the Fraxtil dataset choreography appears to be easy compared to the multi-author ITG dataset. 
We believe this owes to the distinctiveness of author styles. 
Because we have so many step charts for Fraxtil, 
the network is able to closely mimic his patterns.
While the ITG dataset contains multiple charts per author, 
none are so prolific as Fraxtil. 

We released a public demo\footnote{\textbf{\texttt{http://deepx.ucsd.edu/ddc}}} using our most promising models 
as measured by our quantitative evaluation.
Players upload an audio file, select a difficulty rating and receive a step chart for use in the StepMania DDR simulator. 
Our demo produces a step chart for a $3$ minute song in about $5$ seconds using an NVIDIA Tesla K40c GPU. 
At time of writing, 
$220$ players have produced $1370$ step charts with the demo. 
We also solicited feedback, on a scale of $1$-$5$, for player ``satisfaction'' with the demo results. 
The $22$ respondents reported an average satisfaction of $3.87$.

A promising direction 
for future work
is to make the selection algorithm
audio-aware.
We know qualitatively that elements in the ground truth choreography tend to coincide with specific musical events:
jumps are used to emphasize accents in a rhythm;
freezes are used to transition from regions of high rhythmic intensity to more ambient sections.

DDR choreography might also benefit 
from an end-to-end approach,
in which a model simultaneously places steps and selects them.
The primary obstacle here is data sparsity at any sufficiently high feature rate.
At $100 \mathit{Hz}$, about $97\%$ of labels are null.
So in $100$ time-steps of unrolling, an
RNN might only encounter $3$ ground truth steps.

We demonstrate that step selection methods are improved by incorporating $\Delta$-beat and beat phase features,
however our current pipeline does not produce this information.
In lieu of manual tempo input,
we are restricted to using $\Delta$-time features 
when executing our pipeline on unseen recordings.
If we trained a model to detect beat phase,
we would be able to use these features for step selection.

\section{Related Work}
\label{sec:related}
Several academic papers address DDR.
These include anthropological studies
\citep{hoysniemi2006international,behrenshausen2007toward}
and two papers that describe approaches to automated choreography.
The first, called \emph{Dancing Monkeys}, 
uses rule-based methods for both step placement 
and step selection~\citep{okeeffe2003dancing}.
The second employs genetic algorithms for step selection, 
optimizing an ad-hoc fitness function~\citep{nogaj2005genetic}.
Neither establishes reproducible evaluation methodology or
learns the semantics of steps from data.

Our step placement task closely resembles the classic problem of musical onset detection \citep{bello2005tutorial,dixon2006revisited}.
Several onset detection papers investigate modern deep learning methodology.
\citet{eyben2010universal} employ bidirectional LSTMs (BLSTMs) for onset detection;
\citet{marchi2014multi} improve upon this work,
developing a rich multi-resolution feature representation;
\citet{schluter2014improved} demonstrate a CNN-based approach (against which we compare) that performs competitively with the prior BLSTM work.
Neural networks are widely used on a range of other MIR tasks, including musical chord detection \citep{humphrey2012rethinking,boulanger2013audio} and boundary detection \citep{ullrich2014boundary}, another transient audio phenomenon.

Our step selection problem resembles the classic
natural language processing 
task of statistical language modeling.
Classical methods, which we consider, include $n$-gram distributions \citep{chen1998empirical,rosenfeld2000two}.
\citet{bengio2003neural} demonstrate an approach to language modeling using neural networks with fixed-length context.
More recently, RNNs have demonstrated superior performance to fixed-window approaches \citep{mikolov2010recurrent,sundermeyer2012lstm,sutskever2011generating}.
LSTMs are also capable of modeling language at the character level \citep{karpathy2015visualizing, kim2016character}.
While a thorough explanation of modern RNNs exceeds the scope of this paper, we point to two comprehensive reviews of the literature \citep{lipton2015critical,greff2016lstm}.
Several papers investigate neural networks for single-note melody generation \citep{bharucha1989modeling,eck2002first, chu2016song,hadjeres2016deepbach} and polyphonic melody generation \citep{boulanger2012modeling}.

Learning to choreograph requires predicting both the \emph{timing} and the \emph{type} of events in relation to a piece of music.
In that respect, our task is similar to audio sequence transduction tasks, such as musical transcription and speech recognition.
RNNs currently yield state-of-the-art performance for musical transcription \citep{bock2012polyphonic, boulanger2013high,sigtia2016end}.
RNNs are widely used for speech recognition \citep{graves2014towards, graves2006connectionist, graves2013speech, sainath2015learning}, and the state-of-the-art method \citep{amodei2015deep} combines convolutional and recurrent networks.
While our work is methodologically similar, 
it differs from the above in that we consider an entirely different application.

\section{Conclusions}
\label{sec:conclusion}
By combining insights from musical onset detection and statistical language modeling,
we have designed and evaluated a number of deep learning methods
for \emph{learning to choreograph}.
We have introduced standardized datasets 
and reproducible evaluation methodology
in the hope of encouraging wider investigation 
into this and related problems.
We emphasize that the sheer volume of available step charts
presents a rare opportunity for MIR:
access to 
large amounts of high-quality annotated data.
This data could help to spur innovation 
for several MIR tasks,
including onset detection, beat tracking, and tempo detection.

\section*{Acknowledgements}
The authors would like to thank Jeff Donahue, Shlomo Dubnov, Jennifer Hsu, Mohsen Malmir, Miller Puckette, Adith Swaminathan and Sharad Vikram for their helpful feedback on this work.
This work used the Extreme Science and Engineering Discovery Environment (XSEDE)~\citep{towns2014xsede}, which is supported by National Science Foundation grant number ACI-1548562. 
GPUs used for this research were graciously donated by the NVIDIA Corporation.

\bibliography{ddc}
\bibliographystyle{icml2017}

\end{document}